\documentclass{article}

\usepackage[preprint]{neurips_2025}

\usepackage[utf8]{inputenc} 
\usepackage[T1]{fontenc}    
\usepackage{hyperref}       
\usepackage{url}            
\usepackage{booktabs}       
\usepackage{amsfonts}       
\usepackage{nicefrac}       
\usepackage{microtype}      
\usepackage[table,xcdraw]{xcolor}         

\usepackage{amssymb}            
\usepackage{mathtools}          
\usepackage{mathrsfs}           
\usepackage{graphicx}           
\usepackage{subcaption}         
\usepackage[space]{grffile}     
\usepackage{url}                
\usepackage{lipsum}             
\usepackage{multirow}
\usepackage{amsmath}
\usepackage{caption}

\usepackage{svg}
\usepackage{booktabs}
\usepackage{algorithm}
\usepackage{algpseudocode}
\usepackage{wrapfig}

\title{FlowQ: Energy-Guided Flow Policies for Offline Reinforcement Learning}

\author{Marvin Alles$^{1,2\thanks{Corresponding author.}}$
\quad Nutan Chen$^{3}$ 
\quad Patrick van der Smagt$^{3, 4}$ 
\quad Botond Cseke$^{1}$  \\
$^1$Machine Learning Research Lab, Volkswagen Group \quad
$^2$Technical University of Munich \\
$^3$Foundation Robotics Labs, Munich \quad
$^4$Eötvös Loránd University Budapest \\
\texttt{\{marvin.alles, botond.cseke\}@volkswagen.de}
}

\begin{document}

\maketitle

\begin{abstract}
The use of guidance to steer sampling toward desired outcomes has been widely explored within diffusion models, especially in applications such as image and trajectory generation. However, incorporating guidance during training remains relatively underexplored. In this work, we introduce energy-guided flow matching, a novel approach that enhances the training of flow models and eliminates the need for guidance at inference time. We learn a conditional velocity field corresponding to the flow policy by approximating an energy-guided probability path as a Gaussian path. Learning guided trajectories is appealing for tasks where the target distribution is defined by a combination of data and an energy function, as in reinforcement learning.
Diffusion-based policies have recently attracted attention for their expressive power and ability to capture multi-modal action distributions. Typically, these policies are optimized using weighted objectives or by back-propagating gradients through actions sampled by the policy. As an alternative, we propose FlowQ, an offline reinforcement learning algorithm based on energy-guided flow matching. Our method achieves competitive performance while the policy training time is constant in the number of flow sampling steps.
\end{abstract}

\section{Introduction}
\label{sec:introduction}

The goal of offline reinforcement learning is to learn policies entirely from pre-collected, static datasets. This is different from the standard online reinforcement learning setting, where an agent interacts with the environment, gathers new data, and receives corrective feedback through these interactions. The offline setting is particularly appealing for applications such as robotics and autonomous driving, where safe interaction is difficult and costly. However, the absence of corrective feedback introduces a major challenge: distributional shift. The state-action distribution in the offline dataset, driven by the behavior policy, differs from the distribution induced by the learned policy.
As a result, standard online reinforcement learning methods can overestimate the value of out-of-distribution actions, weakening overall performance and causing policy collapse. To address this, offline reinforcement learning methods use various approaches to mitigate value overestimation and improve policy learning.

Most offline reinforcement learning methods constrain the learned policy to stay close to the behavior policy to mitigate the problem of value overestimation \citep{cql, iql, td3bc}. Others take a model-based approach, first learning the environment dynamics before optimizing the policy \citep{MOPO, Lompo, alles2025constrainedlatentactionpolicies}, or formulate offline reinforcement learning as a trajectory prediction problem \citep{TT, Diffuser}. With the success of diffusion models in various fields \citep{sohldickstein2015deepunsupervisedlearningusing, ho2020denoisingdiffusionprobabilisticmodels}, diffusion policies have been introduced, replacing the conventional Gaussian policy with a diffusion model. As offline datasets are often collected from multiple policies they exhibit strong multimodalities and dependencies across action dimensions. As a result, they cannot be effectively modeled by diagonal Gaussian policies but can be better captured using diffusion models \citep{wang2023diffusionpoliciesexpressivepolicy, shafiullah2022behaviortransformerscloningk}. These approaches belong to the first category of offline reinforcement learning methods, as they regularize the distance or divergence between the behavior distribution and the learned policy in various ways. The general idea is to maximize the log-likelihood of the diffusion model for samples from the behavior distribution while incorporating Q-learning to improve policy performance \citep{wang2023diffusionpoliciesexpressivepolicy}.

Modeling multimodal and skewed distributions need to learn the behaviour distribution with diffusion models typically requires many diffusion steps. This makes incorporating Q-learning in training costly since it requires backpropagating through the samples used for Q-function evaluation. To mitigate the latter problem, \citet{wang2023diffusionpoliciesexpressivepolicy} use limited diffusion steps to 5, while \citet{kang2023efficientdiffusionpoliciesoffline} proposes a single step  approximation for sample generation when evaluating the Q-function. An alternative line of work replaces direct optimization with weighted regression \citep{zhang2025energyweighted} or applies guidance w.r.t the Q-function during inference \citep{lu2023contrastiveenergypredictionexact, zheng2023guidedflowsgenerativemodeling}. Our method, FlowQ, aims to minimize the need for action sampling from the model during training, making it scalable with the number of generation steps. It is built on the idea of incorporating guidance during training to directly learn a policy of the form $\pi_\theta(a\vert s) \propto \pi_{\beta}(a\vert s)\exp(Q(s, a))$.

Flow Matching has recently gained popularity as a method for training residual models, relying on ordinary differential equations (ODEs) rather than stochastic differential equations (SDEs). It simplifies training while also demonstrating competitive performance on a variety of problems \citep{esser2024scalingrectifiedflowtransformers, black2024pi0visionlanguageactionflowmodel, vyas2023audioboxunifiedaudiogeneration}. Building on this, we propose FlowQ, a Q-function guided flow matching approach that learns a time-dependent velocity field corresponding to a probability path based on the optimal policy $\pi_\theta(a\vert s) \propto \pi_{\beta}(a\vert s)\exp(Q(s, a))$. To do this, we propose a conditional probability path that in addition to data also uses the Q-function and a corresponding Gaussian path approximation that allows us to sample and compute the velocity field. Unlike other methods, FlowQ does not require additional sampling from the model during training, except for learning the Q-function, which is common in all Q-learning-based approaches. This makes FlowQ scalable to a large number of generation steps. Furthermore, FlowQ does not introduce extra overhead during inference, unlike diffusion methods that rely on guidance at this stage. Overall, FlowQ seamlessly integrates into the simulation-free flow matching framework.

\begin{figure}[t]
    \begin{center}
        \includegraphics[width=0.9\columnwidth]{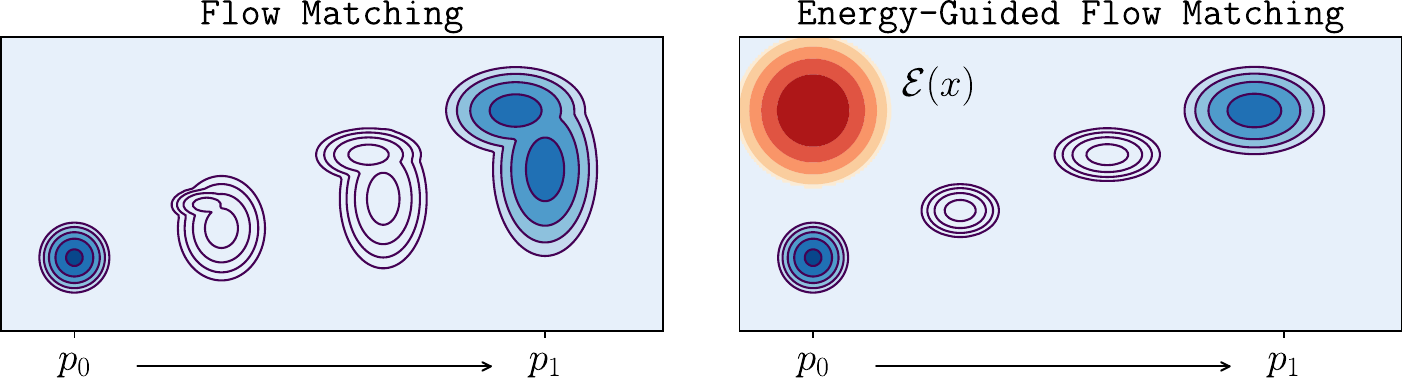}
    \end{center}
    \caption{\small \textit{Energy-Guided Flow Matching}. Flow matching learns a flow to map from a known source distribution $p_0(x_0)$ to the data distribution $p_1(x_1)$. 
    In contrast, energy-guided flow matching seeks to learn a transformation from a known source distribution $p_0(x_0)$ to an energy-weighted distribution $\hat p_1(x_1) \propto p_1(x_1) \exp \{- \lambda \,\mathcal{E}(x_1) \}$ instead. In the minimal example shown in the figure, the energy function $\mathcal{E}$ guides the flow toward generating the upper mode of the multimodal data distribution.}
    \label{fig:egfm}
\end{figure}

We summarize our contributions as follows:
\begin{itemize}
    \item[(1)] We introduce energy-guided flow matching, a method to learn an approximation of $p(x) \propto q(x)\exp(-\lambda\, \mathcal{E}(x))$ where $q(x)$ is given by a set of samples (empirical distribution) and we assume that the derivatives of  $\mathcal{E}(x)$ are available.
    \item[(2)] We propose FlowQ, an offline reinforcement learning method based on energy-guided flow matching with $Q$ as an energy function. The training time scales well w.r.t. the number of flow generation steps because we do not need to backpropagate gradients through actions sampled from the flow policy.
    \item[(3)] We evaluate FlowQ on the D4RL benchmark and compare it to other state of the art offline reinforcement learning methods.
\end{itemize}

\section{Preliminaries}
\label{sec:preliminaries}
\textbf{Offline Reinforcement Learning.}
In reinforcement learning, an environment is usually defined as a Markov decision process (MDP) by $M = (\mathcal{S},\mathcal{A},T,R,\gamma)$ with $\mathcal{S}$ as state space, $\mathcal{A}$ as action space, $s \in \mathcal{S}$ as state, $a \in \mathcal{A}$ as action, $T : \mathcal{S} \times \mathcal{A} \rightarrow \mathcal{S}$ as transition function, $R : \mathcal{S} \rightarrow R$ as reward function and $\gamma \in (0, 1]$ as discount factor. The goal is to
learn a policy $\pi : \mathcal{S} \rightarrow \mathcal{A}$ that maximizes the expected discounted sum of rewards $\mathbb{E}[\sum^T_{t=1} \gamma^t r_t]$ \citep{sutton1998introduction}. In online reinforcement learning an agent interacts with the environment to optimize its policy. In contrast, offline reinforcement learning requires the agent to optimize its policy using a static dataset $\mathcal{D}=\{s_t, a_t, s_{t+1}, r_t\}$, which was collected beforehand by an unknown behavior policy $\pi_\beta$.
Among the various approaches for learning a policy in this setting, diffusion policies have gained popularity due to their expressiveness and ability to capture multi-modal action distributions, which are often common in static datasets \citep{kang2023efficientdiffusionpoliciesoffline,wang2023diffusionpoliciesexpressivepolicy, lu2023contrastiveenergypredictionexact}.

\textbf{Flow Matching.}
Flow matching is a method for generative modeling, specifically for training residual models \citep{lipman2023flowmatchinggenerativemodeling}. 
Unlike standard diffusion models, which model dynamics via stochastic differential equations (SDEs), flow matching uses ordinary differential equations (ODEs) to describe a deterministic transformation (no additive noise) of a random sample from an initial distribution, thereby simplifying training and linking naturally to continuous normalizing flows: Flow matching allows the training of continuous normalizing flows in a simulation-free manner.

The goal of flow matching is to learn a model to transform samples from a known source distribution $p_0$ to a data distribution $p_1$, by learnig an ODE model. 
Let $x_0 \sim p_0(x_1)$ and let $x_t=\psi_t(x_0)$ with $\psi_t: [0, 1] \times \mathbb{R}^d \rightarrow \mathbb{R}^d$ be defined by an ODE model $u_t(\cdot)$ via
\begin{align}
    \frac{d}{dt}\psi_t(x_0)=u_t(\psi_t(x_0)).
    \label{eq:ode}
\end{align}
The function $u_t: [0, 1] \times \mathbb{R}^d \rightarrow \mathbb{R}^d$ is called the velocity field and the resulting solutions $\psi_t(x_0), t\in [0,1], x_0 \sim p_0(x_0)$ (flow map) transform $p_0(x_0)$ into $p_t(x_t)$ with $x_t = \psi_t(x_0) \sim p_t(x_t)$. 

To learn a parametric model of the velocity field $u^\theta_t$, (conditional) flow matching minimizes the objective 
\begin{align*}
    \mathcal{L}_{\scriptsize{\mathrm{CFM}}}(\theta)= \mathbb{E}_{t \sim \mathcal{U}[0, 1], x_t \sim p_t, x_1\sim q}
    \left[
    \vert\!\vert 
    u^{\theta}_t(x_t) - u_t(x_t \vert x_1) 
    \vert\!\vert^2 
    \right],
\end{align*}
where the target velocity field $u_t(x_t \vert x_1)$ is conditioned on a sample from the data distribution $x_1\sim q$. The conditional distribution $p_t(x_t \vert x_1)$, called the conditional probability path, is typically modeled as the conditional Gaussian distribution
  $  p_t(x_t \vert x_1)=\mathcal{N}(x_t; \alpha(t, x_1),\sigma(t)^2I) \text{,}$
resulting in the flow representation
$
    x_t =  \psi(x_t) = \alpha(t, x_1) + \sigma(t)\epsilon\text{,}
$
and a corresponding conditional velocity field
$
    u_t(x_t \vert x_1) = \dot \alpha(t) + \frac{\dot\sigma(t)}{{\sigma(t)}}[x_t - \alpha(t)] \text{.}
$
The marginal probability path $p_t(x)$ can be obtained by integrating over $x_1$
$
    p_t(x)=\int  p_t(x \vert x_1) p(x_1) dx_1  \text{.}
$
A~popular instance of a Gaussian path, which can be derived as the solution to an Optimal Transport problem, is the linear path
$
    p_t(x_t \vert x_1)=\mathcal{N}(x_t; tx_1,(1-t)^2I) \text{,}
$
which allows us to express the corresponding flow as
$
    x_t =  \psi_t(x_0) = tx_1 + (1-t)x_0\text{,}
$
and consequently, the target velocity simplifies to
$
    u_t(x_t \vert x_1) = (x_1 - x_t)/(1-t) = x_1 - x_0(x_t, x_1) \text{.}
$
Once the velocity field $u^\theta_t$ is learned, new samples can be generated by numerically solving the ODE defined in Equation~\eqref{eq:ode}.

\section{Related Work}
\label{sec:related_work}
\textbf{Offline Reinforcement Learning.}
Most offline reinforcement learning methods are based on the concept of  regularizing the policy. This is oftentimes achieved by incorporating some form of behavior cloning. Broadly, these approaches can be categorized into direct policy optimization and weighted regression.
A well-known method in the first category is TD3+BC, which extends TD3 by adding behavior cloning to ensure that policy-generated actions remain close to the dataset's action distribution \citep{td3bc}. Similarly, PLAS and BCQ \citep{PLAS, BCQ} employ a conditional variational autoencoder to estimate the behavior distribution and constrain the policy accordingly. Beyond these, numerous other approaches exist. For instance, CQL \citep{cql} enforces conservatism by optimizing a lower bound on value estimates, thus mitigating value overestimation on of out-of-distribution actions.
The second category includes methods like AWR and CRR \citep{peng2021advantageweighted, crr}, which weight a behavior cloning term by using the Q-function to prioritize high-value actions. Another popular approach is IQL \citep{iql}, which uses expectile regression to implicitly constrain the policy to avoid out-of-distribution actions.

\textbf{Diffusion and Flow Matching in Offline Reinforcement Learning.}
Recent advancements in diffusion models \citep{ho2020denoisingdiffusionprobabilisticmodels, sohldickstein2015deepunsupervisedlearningusing} and flow matching \citep{lipman2023flowmatchinggenerativemodeling} have led to the adoption of residual models in reinforcement learning: \citep{lu2023syntheticexperiencereplay, jackson2024policyguideddiffusion} use a diffusion model to generate synthetic data to augment the dataset. Similarly, \citet{park2025flowQlearning} presents an algorithm where a flow model is used to distill the offline data and a TD3-BC-like update (policy improvement) is used to learn an MLP policy. In addition to these methods, a second category of approaches exists, which integrates diffusion and flow models as policies:
DiffusionQL \citep{wang2023diffusionpoliciesexpressivepolicy} integrates a diffusion-based policy into a TD3+BC-like framework, while EDP \citep{kang2023efficientdiffusionpoliciesoffline} improves efficiency by introducing a single-step action approximation, among other optimizations. Another line of research uses guidance during inference to direct policies toward high-value actions. Guided Flows \citep{zheng2023guidedflowsgenerativemodeling} transfers the concept of guidance during inference from diffusion models to flow matching. CEP \citep{lu2023contrastiveenergypredictionexact} learns a time-dependent energy function to better approximate the posterior distribution and DAC \citep{fang2025diffusionactorcriticformulatingconstrained} integrates soft Q-guidance directly into policy learning. Additionally, methods such as QIPO, QVPO, and certain EDP variants \citep{zhang2025energyweighted, kang2023efficientdiffusionpoliciesoffline} fall under the category of weighted regression approaches, as they primarily rely on importance-weighted learning to guide policy updates.

\section{Energy-Guided Flow Policies} 

\subsection{Energy-Guided Flow Matching}
\label{sec:flowq}
In contrast to the conventional generative modeling paradigm, which seeks to learn a transformation from a known source distribution $p_0(x_0)$ to the data distribution $p_1(x_1)$, we propose an alternative formulation called \textit{energy-guided flow matching}. In our framework, samples from the source distribution $p_0(x_0) = p(x_0)$ are not mapped directly to the target distribution $p_1(x_1)$ but to an energy-weighted distribution 
\begin{align*}
    \hat p_1(x_1) \propto p_1(x_1) \exp \{- \lambda \,\mathcal{E}(x_1)\}
\end{align*}
with $\mathcal{E}: \mathcal{R}^d \rightarrow \mathcal{R}$ as energy function and $\lambda \in \mathcal{R}^+$
as a scaling parameter that modulates the strength of the energy guidance. We provide a visualization in Figure \ref{fig:egfm}, where energy-guided flow matching is approximating the posterior, given samples of the data distribution and an energy function $\mathcal{E}(x)$. Unlike previous approaches that define similar energy-guided distributions for guided sampling during inference \citep{lu2023contrastiveenergypredictionexact, zheng2023guidedflowsgenerativemodeling}, our objective is to learn a parametric velocity field $\hat u^\theta_t$ that generates the energy-guided marginal probability path
$
\hat p_t(x_t) = \int \hat p_t(x_t \vert x_1) \hat p_1(x_1) dx_1 \text{.}
$

We propose that at time $t$ the energy-guided marginal probability path is given by the approximation
\begin{align*}
    \hat p_t(x_t) \propto p_t(x_t) \exp \{- \lambda(t)\, \mathcal{E}(x_t)\} \text{,}
\end{align*}
where the energy function $\mathcal{E}(x_1)$ is independent of time, but scaled by a time-dependent factor~$\lambda(t)$. Building on the standard flow matching assumption of Gaussian paths, which implies that $p_0(\cdot \vert x_1) = p(\cdot)$, we choose
$
    \lambda(t) = h(t) \lambda \text{, with } h(0)=0 \text{.}
$
This is different to the formulation in \citep{lu2023contrastiveenergypredictionexact}, which defines an intermediate energy function dependent on $t$. 
We derive the corresponding conditional probability path as
\begin{align*}
    \hat p_t(x_t \vert x_1) =  p_t(x_t \vert x_1) \frac{\exp \{- \lambda(t)\,\mathcal{E}(x_t)\} Z_1}{\exp \{- \lambda(1) \,\mathcal{E}(x_1) \} Z_t} \text{,}
\end{align*}
which cannot be expressed in closed form (Appendix \ref{eq:appendix_cond_prob}). Since the normalization factors $Z_1$, $Z_2$ and $\exp \{- \lambda(1)\, \mathcal{E}(x_1)\}$ are independent of $x_t$ and only affect the overall scale of the density (not its mean or variance), we can simplify the expression to
\begin{align*}
    \hat p_t(x_t \vert x_1) \propto  p_t(x_t \vert x_1) \exp \{- \lambda(t)\, \mathcal{E}(x_t)\}\text{.}
\end{align*}

To derive a tractable expression for the conditional probability path, we choose
$
    p_t(x_t \vert x_1) = \mathcal{N}(x_t; tx_1, (1-t)^2 I)
$
and approximate the energy function $\mathcal{E}(x)$ using a first-order Taylor expansion around the mean of $p_t(x_t \vert x_1)$:
\begin{align*}
    \mathcal{E}(x_t) \approx \mathcal{E}(tx_1) + \nabla\mathcal{E}(tx_1)^T(x_t-tx_1) \text{.}
\end{align*}
The full derivation can be found in Appendix \ref{eq:appendix_taylor}. Under this linear approximation the conditional probability path is Gaussian
$
    \hat p_t(x_t \vert x_1) = \mathcal{N}(x_t; \alpha_c(t, x_1), \sigma_c(t)^2 I)
$
where the variance remains unchanged,
$
    \sigma_c(t)^2 = (1-t)^2
$
and the mean is shifted according to
$
    \alpha_c(t, x_1) = tx_1 -  (1-t)^2 \lambda(t) \nabla\mathcal{E}(t x_1) \text{.}
$
The conditional velocity field and conditional flow are then defined through the standard flow matching expressions for Gaussian paths (Appendix \ref{eq:target_ut}), which are given by
\begin{align*}
    \hat u_t(x_t \vert x_1) &= \dot \alpha_c(t, x_1) + \frac{\dot\sigma_c(t)}{{\sigma_c(t)}}[x_t - \alpha_c(t, x_1)] \\
    &= \frac{x_1-x_t}{1-t} + (1-t) [\lambda(t) -  (1-t) \dot\lambda(t)]\nabla\mathcal{E}(tx_1) -  (1-t)^2 \lambda(t) \nabla^2\mathcal{E}(tx_1)x_1
\end{align*}
and $x_t$ can be sampled via
$
x_t(x_1, \epsilon ) = tx_1 -  (1-t)^2 \lambda(t)\nabla\mathcal{E}(tx_1) + (1-t)\epsilon
$
with
$
\epsilon \sim \mathcal{N}(0, I).
$

\begin{figure}[ht]
\centering
    \begin{subfigure}{0.8\columnwidth}
        \begin{center}
            \includegraphics[width=1.0\columnwidth]{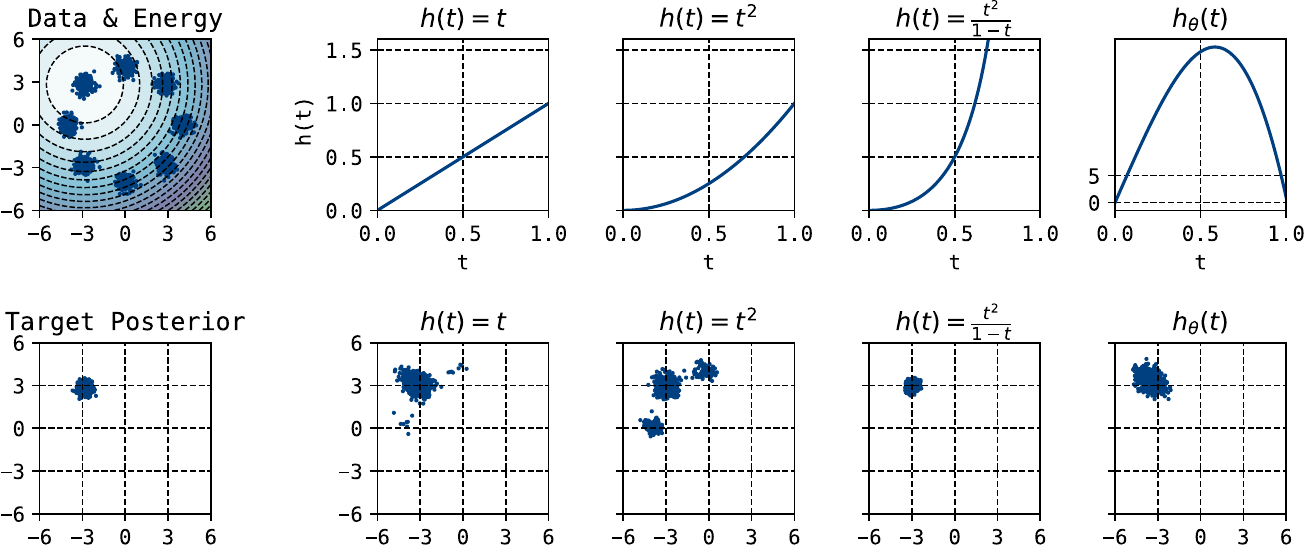}
        \end{center}
        \subcaption{The energy function $\mathcal{E}(x)$ is aligned with the center of the cluster in the left corner.}
    \end{subfigure}\vspace{0.02\columnwidth}
    \begin{subfigure}{0.8\columnwidth}
        \begin{center}
            \includegraphics[width=1.0\columnwidth]{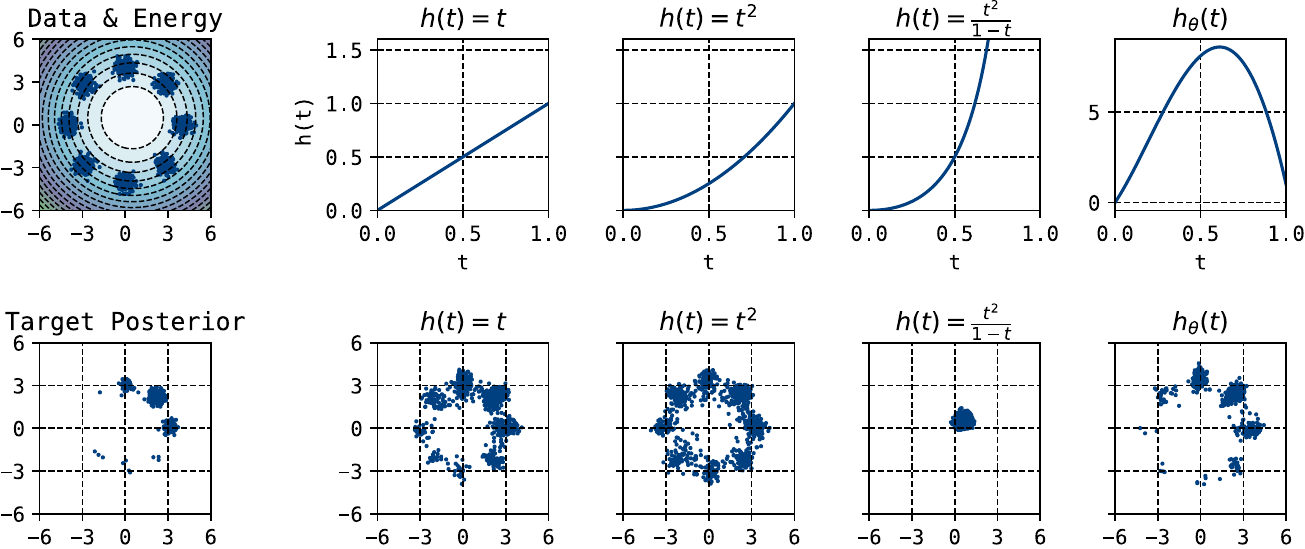}
        \end{center}
        \subcaption{The energy function $\mathcal{E}(x)$ is centered at $[1, 1]$, which is outside of the data distribution.}
    \end{subfigure}

    \caption{\small Visualization of energy-guided flow matching. Given samples of the data distribution and an energy function, energy-guided flow matching approximates the posterior. We compare fixed energy scaling schedules $t$, $t^2$, $\frac{t^2}{1-t}$, as well as a learnable schedule $h_{\theta}(t)$.}
    \label{fig:h_study}
\end{figure}

\textbf{Time-depedent scaling.} As the choice of $h(t)$ impacts the performance and is dependent on the task we explore the following variants: fixed schedules $t$, $t^2$, $\frac{t^2}{1-t}$, as well as a learnable schedule $h_{\theta}(t)$. Specifically, we aim to analyze how our choices affects performance, considering scenarios where the energy maximum either overlaps with the data distribution or lies outside of it. Figure \ref{fig:h_study} shows the results on a regression task with data sampled from a Gaussian mixture model. In our analysis, the first scenario uses an energy function centered on one of the clusters, while the second uses an energy function centered at $[1,1]$, which falls outside the data distribution. Each variant of $h(t)$ influences differently how much weight is assigned to the energy distribution relative to the data distribution as a function of the sampling step $t$. For example, the schedule $t$ assigns more weight to the energy compared to $t^2$, and $t^2 / (1-t)$ becomes unbounded as $t$ approaches $1$, thereby emphasizing the energy maximum even more. We implement the learnable schedule as $h_{\theta}(t)=t + t(1 - t)f_{\theta}(t)$ which guarantees $h(0)=0$ and $h(1)=1$, with $f(t)$ as a multilayer perceptron (MLP). Finally, we rescale $h_{\theta}(t)$ by $\frac{1}{\mathbb{E}_{x\sim \mathcal{D}} \vert \mathcal{E}(x)\vert}$ to eliminate the influence of different energy scales. Overall, when the energy maximum is located outside the data distribution, the learnable schedule most closely approximates the target posterior. Moreover, when the energy maximum overlaps with the data distribution, the learnable schedule achieves the second-best performance after the maximum-seeking variant $\frac{t^2}{1-t}$. Nevertheless, it remains an approximation of the posterior, leaving room for further improvement. 

\textbf{Comparison to Energy-Weighted Flow Matching.}
In \citep{lu2023contrastiveenergypredictionexact} the authors show that the marginal probability flow for a model with target distribution $\hat{p}_1(x_1)$ and conditional probability path $p_t(x_t \vert x_1)$ satisfies $\hat{p}_t(x_t) \propto p_t(x_t) \exp\{ -\mathcal{E}_t(x_t)\}$ with $ \mathcal{E}_t(x_t) = - \log\mathbb{E}_{p_t(x_1 \vert x_t) }[ \exp\{ \lambda \mathcal{E}(x_1)\}]$, thus providing the guided flow via $\nabla_x \hat{p}_t(x_t) =  \nabla_x p_t(x_t) - \nabla_x \mathcal{E}_t(x_t)$.
This form of $\mathcal{E}_t(x_t)$ presents several challenges both because of the intractability of the integrand and because of the averaging distribution. For this reason, a novel method (contrastive energy prediction) is proposed to simultaneously learn a parametric version of $\mathcal{E}_t(\cdot)$ to follow $Q(\cdot)$. The authors in \citep{zhang2025energyweighted} use the form of $\mathcal{E}_t(x_t)$ to propose an energy-weighted flow matching and a corresponding conditional energy-weighted flow matching objective. The latter can be viewed as a conditional flow matching objective where each loss term associated to a sample $x_1^i$ is weighted by its normalized weight w.r.t. the distribution $\hat{p}_1(x)$, that is, by $\exp\{-\lambda \mathcal{E}(x_1^i) \} / \mathbb{E}_{p_1(x_1)}[\exp\{-\lambda \mathcal{E}(x_1)\}]$. In comparison, we propose an approximation of the conditional probability path and the corresponding velocity field assuming $\mathcal{E}_t(x_t) \approx \lambda(t) \mathcal{E}(x_t)$ and using a first-order expansion inspired by~\citet{sohldickstein2015deepunsupervisedlearningusing}.


\subsection{Offline reinforcement learning}

The goal of Offline RL is to learn a policy, which maximizes the expected discounted sum of rewards given an already collected dataset of $\{s_t, a_t, s_{t+1}, r_t\}$ tuples. Methods such as TD3+BC \citep{td3bc}, CQL \citep{cql}, or DiffusionQL \citep{wang2023diffusionpoliciesexpressivepolicy} try to achieve the objective by regularised policy optimization. However, this approach is computationally expensive when using diffusion or flow policy models because  it requires back-propagating gradients through actions sampled from the policy.
An alternative is to use likelihood-based methods, which formulate the learning objective as
\begin{align}
\mathcal{L}_{\pi}(\theta) &= \mathbb{E}_{s_t \sim \mathcal{D}}
\left[
\mathrm{D}_{\mathrm{KL}}
\left[
Z^{-1}
e^{Q(s_t, a_t)}
\pi_{\beta}(a_t \vert s_t)
\,\big|\!\big|\,
\pi_{\theta}(a_t \vert s_t) 
\right]
\right]
\\
&= - \mathbb{E}_{s_t \sim \mathcal{D}}\mathbb{E}_{\pi_{\beta}(a_t \vert s_t)\exp(Q(s_t, a_t))}
\left[
\log \pi_{\theta}(a_t \vert s_t)
\right]
+ \mathrm{const}\text{.}
\label{EqnRlObj}
\end{align}
Among them are AWR \citep{peng2021advantageweighted} and IQL \citep{iql} which use weighted regression to learn a policy. Similarly, weighted diffusion-based and flow-based methods such as EDP \citep{kang2023efficientdiffusionpoliciesoffline} and QIPO \citep{zhang2025energyweighted} exist, but they can be expensive to compute when sampling from the policy is required during training. Overall, our goal is to learn a policy in a simulation-free manner, following the principles of flow matching. Therefore sampling from the policy during training is an option we do not consider. Directly optimizing the objective via flow matching is infeasible because we cannot sample from $\pi_{\beta}(a_t \vert s_t)\exp(Q(s_t, a_t))$. However, we do have samples from $\pi_{\beta}(a_t \vert s_t)$ and can evaluate $\exp(Q(s_t, a_t))$.
Hence, we propose \textit{FlowQ} as an approach based on energy-guided flow matching that aims to learn a policy which satisfies
$
    \pi(a_t \vert s_t) \propto \pi_{\beta}(a_t \vert s_t) \exp(Q(s_t, a_t)) \text{.}
$
The Q-function is learned in a conventional way by minimizing the Bellman error. The complete method is outlined in Algorithm~\ref{AlgoFlowQ}.

\begin{algorithm}[ht]
\centering
\caption{\small FlowQ, an energy-guided flow matching based offline RL algorithm.}
\label{alg:flowq}
\begin{algorithmic}
\footnotesize
\State Given dataset $\mathcal{D}$, critic $Q_{\phi}$ with parameters $\phi_{i=1,2}$ and flow model $u_t^{\theta}$, and target networks $\phi'_{i=1,2}$ and $u_t^{\theta'}$.
\While{not converged}
\State Sample batch $ \mathcal{B}=\{(s, a, r, s')_i\}_{i=1}^{b}\sim \mathcal{D}$
\State ~
\State $\triangleright$ \textbf{Train Critic}
\ForAll {$(s, a, r, s') \in \mathcal{B}$}
\State $a' \sim \pi_{\theta'}(a'\vert s')$ via 
$a' \gets \text{ODE}(u^{\theta'}(\cdot, s'), a'_0),\: a'_0 \sim \mathcal{N}(0,1)$
\State $\hat Q \leftarrow  \mathbb{E}_{a'\sim \pi_{\theta'}}(r + \gamma \min_{i=1,2} Q_{\phi'_i}(s', a'))$
\State Update $\phi_{i=1,2}$ by minimizing step on $L(\phi_i) = ( Q_{\phi_i}(s, a) - \hat Q)^2$
\EndFor  
\State ~
\State $\triangleright$ \textbf{Train Flow Policy}
\ForAll {$(s, a, r, s') \in \mathcal{B}$}
\State $a_1 \leftarrow a$
\State $\epsilon \sim \mathcal{N}(0,I)$, $t \sim \mathcal{U}(0, 1)$
\State $a_t(a_1, s, \epsilon ) \leftarrow ta_1 -  (1-t)^2 \lambda(t)\nabla Q(s, ta_1) + (1-t)\epsilon$
\State $\hat u_t(a_t \vert a_1, s) \leftarrow \dot \alpha_c(t, a_1,s) + \frac{\dot\sigma_c(t)}{{\sigma_c(t)}}[a_t - \alpha_c(t, a_1,s)]$
\State Update $\theta$ by minimizing step on
$
L(\theta) = 
    \vert\!\vert 
    u^{\theta}_t(a_t, s) - \hat{u}_t(a_t \vert a_1, s) 
    \vert\!\vert^2 
$
\EndFor  
\State ~
\State $\triangleright$ \textbf{Update target networks}
\State $\theta' \leftarrow \rho \theta'+(1-\rho)\theta$
\State $\phi' \leftarrow \rho \phi'+(1-\rho)\phi$

\EndWhile

\end{algorithmic}
\label{AlgoFlowQ}
\end{algorithm}

\section{Experiments}
\label{sec:experiments}
We evaluate FlowQ to address the following key questions: (1) Is energy-weighted flow matching competitive with other methods? (2) How efficient is our method in terms of training time, and how well does it scale with the number of sampling steps? We evaluate the method on $18$ datasets from the D4RL benchmark \citep{D4RL} including tasks from locomotion, navigation and manipulation.

\textbf{Baselines.}
We select a set of relevant baselines to compare against FlowQ. As a simple reference, we include a Behavior Cloning (BC) baseline, which evaluates how well a policy performs without using reward information. For classic offline reinforcement learning methods using Gaussian policies, we compare against IQL \citep{iql}, CQL \citep{cql}, and TD3+BC \citep{td3bc}, reporting their scores from \citep{iql}. As a representative of early diffusion-based policy methods, we include DiffusionQL \citep{wang2023diffusionpoliciesexpressivepolicy}. Additionally, we implement a variant where we replace the diffusion model with a flow-matching model to analyze the effect of these different paradigms, namely (FM)DiffusionQL. EDP \citep{kang2023efficientdiffusionpoliciesoffline} builds on DiffusionQL by introducing action approximation to accelerate policy training. To maintain a fair comparison, we implement it with 20 diffusion steps as in FlowQ and do not include other sampling strategies. We also report results from Guided Flows \citep{zheng2023guidedflowsgenerativemodeling}, which uses Q-function guidance during inference and QIPO, which uses an energy-weighted flow matching objective. For consistency, we ensure that DiffusionQL, (FM)DiffusionQL, EDP, and FlowQ share most implementation details, allowing for a fair comparison. 

\textbf{Experimental Setup.}
We evaluate each method using the \textit{running average at training} protocol proposed by \citet{kang2023efficientdiffusionpoliciesoffline}, which reports the average score of the last 10 evaluations. Specifically, we evaluate each policy every 20000 gradient steps by averaging returns across 20 independent rollouts, and repeat each experiment with five random seeds. 
We implement everything using JAX and provide hyper-parameters for FlowQ in Appendix \ref{appendix:implementation}. For (FM)DiffusionQL and EDP we follow the hyper-parameters and implementation details of \citep{wang2023diffusionpoliciesexpressivepolicy}. Since evaluation protocols vary across offline reinforcement learning publications, we include Table \ref{tab:best_result} in the appendix to report scores measured using the \textit{best result during training} method, as used in \cite{zhang2025energyweighted} for QIPO. We further add the results for DiffusionQL reported in  \cite{wang2023diffusionpoliciesexpressivepolicy} for \textit{online model selection}.

\subsection{Performance on the D4RL Benchmark}
We evaluate the methods on nine locomotion datasets, including the medium-expert, medium, and medium-replay variants of the halfcheetah, walker, and hopper environments. Additionally, we include six antmaze navigation environments and two adroit manipulation environments. The averaged scores are provided in Table \ref{tab:d4rl_results}. Across locomotion tasks, FlowQ achieves competitive performance overall, with the exception of halfcheetah-medium-expert and hopper-medium-expert, where its results are slightly lower. Similarly, the results of (FM)DiffusionQL and DiffusionQL are comparable, indicating that flow matching alone does not guarantee better performance but serves as a reasonable alternative. While DiffusionQL is competitive for the antmaze environments as well, (FM)DiffusionQL fails to learn an effective policy for the more challenging antmaze-medium and antmaze-large environments. In contrast, FlowQ achieves the highest performance on the antmaze and the manipulation (pen-human and pen-cloned) environments, demonstrating the benefits of energy-guided flow matching compared to a naive TD3+BC-like framwork as in DiffusionQL, (FM)DiffusionQL and EDP. We additionally show the effectiveness of FlowQ compared to QIPO in Appendix \ref{appendix:implementation}.

\begin{table}[ht]
\resizebox{\textwidth}{!}{ 
    \begin{tabular}{lccccccccc}
    \toprule
    Dataset                   & BC    & IQL   & CQL   & TD3+BC & DiffusionQL\textsuperscript{*} & (FM)DiffusionQL\textsuperscript{*} & EDP\textsuperscript{*}   & Guided Flows & FlowQ(ours)\textsuperscript{*} \\
    halfcheetah-medium        & 42.6  & 47.4  & 44.0  & 48.3   & 47.9        & 47.6            & 48.2  & 49.0        & \textbf{56.8 $\pm$ 0.8}  \\
    halfcheetah-medium-replay & 36.6  & 44.2  & 45.5  & 44.6   & 45.2        & 44.6            & 45.5  & 42.0        & \textbf{50.3 $\pm$ 0.9}  \\
    halfcheetah-medium-expert & 55.2  & 86.7  & 91.6  & 90.7   & 93.4        & 93.0            & 90.6  & \textbf{97.0}          & 89.6 $\pm$ 2.3  \\
    hopper-medium             & 52.9  & 66.3  & 58.5  & 59.3   & 88.0        & 65.7            & 68.7  & 84.0         & \textbf{88.9$ \pm$ 6.4}   \\
    hopper-medium-replay      & 18.1  & 94.7  & 95.0  & 60.9   & 98.7        & 98.3            & 98.1  & 89.0          & \textbf{99.2 $\pm$ 1.4}  \\
    hopper-medium-expert      & 52.5  & 109.6 & 105.4 & 98.0   & \textbf{110.1}       & 109.6           & 108.0 & 105.0        & 103.5 $\pm$ 7.0 \\
    walker2d-medium           & 75.3  & 78.3  & 72.5  & 83.7   & 84.7        & 83.7            & 83.7  & 77.0          & \textbf{84.9 $\pm$ 0.4}  \\
    walker2d-medium-replay    & 26.0  & 73.9  & 77.2  & 81.8   & 87.5        & 86.2            & 87.2  & 78.0          & \textbf{88.5 $\pm$ 2.9}  \\
    walker2d-medium-expert    & 107.5 & 109.6 & 108.8 & \textbf{110.1}  & 109.1       & 109.0           & 109.0 & 94.0          & 109.1 $\pm$ 0.2 \\
    \rowcolor[HTML]{EFEFEF} 
    average                   & 51.9  & 77.0  & 77.6  & 75.3   & 85.0        & 82.0            & 82.1  & 79.4          & \textbf{85.6}   \\
    antmaze-umaze             & 54.6  & 87.5  & 74.0  & 40.2   & 57.0        & 61.0            & 60.0  & -             & \textbf{92.8 $\pm$ 7.9}  \\
    antmaze-umaze-diverse     & 45.6  & 62.2  & \textbf{84}  & 58.0   & 61.5        & 59.8            & 60.2  & -            & 65.2 $\pm$ 13.3  \\
    antmaze-medium-play       & 0.0   & 71.2  & 61.2  & 0.2    & \textbf{84.8}        & 0.0             & 78.8  & -            & \textbf{84.8 $\pm$ 7.9}  \\
    antmaze-medium-diverse    & 0.0   & \textbf{70.0}  & 53.7  & 0.0    & 52.2        & 0.0             & 57.7  & -            & 44.7 $\pm$ 22.8.6  \\
    antmaze-large-play        & 0.0   & 39.6  & 15.8  & 0.0    & 58.0        & 0.0             & 47.8  & -            & \textbf{72.4 $\pm$ 11.5}   \\
    antmaze-large-diverse     & 0.0   & 47.5  & 14.9  & 0.0    & 40.8        & 0.0             & 29.8  & -             &  \textbf{55.8 $\pm$ 16.8}   \\
    \rowcolor[HTML]{EFEFEF} 
    average                   & 16.7  & 63.0  & 50.6  & 16.4   & 59.1        & 30.1            & 55.7  & -             & \textbf{69.3}      \\
    pen-human                 & 63.9  & \textbf{71.5}  & 37.5  & 5.9    & 67.6        & 67.9            & 70.3  & -            & 69.5 $\pm$ 19.0  \\
    pen-cloned                & 37.0  & 37.3  & 39.2  & 17.2   & 60.4        & 68.4            & 60.7  & -            & \textbf{86.1 $\pm$ 20.3}  \\
    \rowcolor[HTML]{EFEFEF} 
    average                   & 50.5  & 54.4  & 38.4  & 11.6   & 64.0        & 68.2            & 65.6  & -            & \textbf{77.8} \\
    \bottomrule
    \\
    \end{tabular}
}
\caption{\small Averaged normalized score on the D4RL benchmark. We report results (mean and standard deviation) using the \textit{running average at training} method as proposed in \citep{kang2023efficientdiffusionpoliciesoffline} and use five seeds for each experiment. The results for methods annotated by * come from our own implementation.}
\label{tab:d4rl_results}
\end{table}

\subsection{Efficiency and Scalability}
To evaluate the efficiency and scalability of FlowQ, we analyse the training time required for updating the policy. In Figure \ref{fig:time} we compare FlowQ to EDP and DiffusionQL over $10^4$ gradient steps. 
We report the training time for one workstation using an NVIDIA GTX 1080. While the training time for FlowQ and EDP remains independent of the number of timesteps, DiffusionQL's training time increases linearly. This is due to the computational overhead of backpropagating gradients through the Q-function, which significantly slows down training.

We note that the training time of the critic usually depends on the number of timesteps, as actions are typically sampled from the model to update the Q-function. However, since this aspect is independent of policy training, we excluded it from our evaluation. Moreover, alternative sampling methods, such as DPM-Solver \citep{lu2022dpmsolverfastodesolver, kang2023efficientdiffusionpoliciesoffline}, can significantly reduce the number of timesteps required for action sampling.

\subsection{Ablation study on $h(t)$ and $\lambda$}
\begin{figure}[ht]
\centering
    \begin{subfigure}{0.22\columnwidth}
        \begin{center}
            \includegraphics[width=0.95\columnwidth]{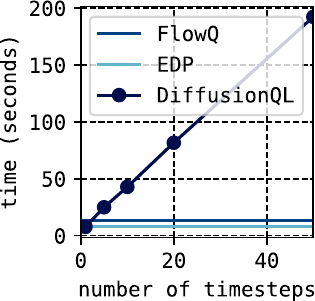}
        \end{center}
        \caption{\small Training time}
    \label{fig:time}
    \end{subfigure}
    \begin{subfigure}{0.72\columnwidth}
        \begin{center}
            \includegraphics[width=0.95\columnwidth]{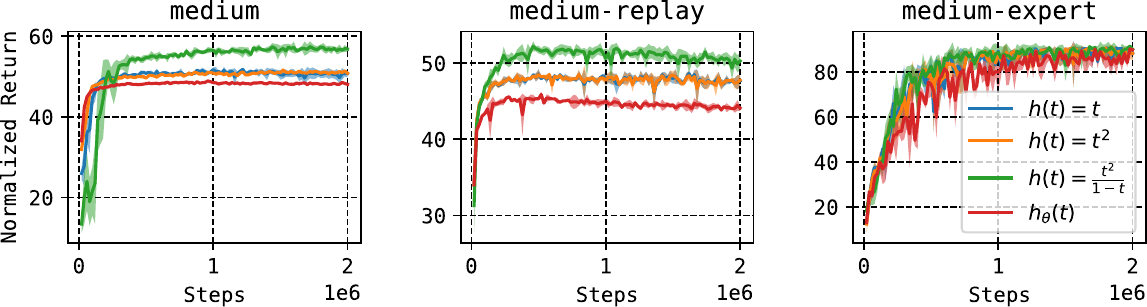}
        \end{center}
        \caption{\small Ablation study on $h(t)$ on the halfcheetah environments.}
    \label{fig:h_ablation}
    \end{subfigure}

    \caption{\small In (a) we plot policy training time in seconds for $10^4$ gradient steps over the number of sampling steps $t$. In (b) we plot the normalized return on halfcheetah environments for different choices of $h(t)$.}
\end{figure}

Since both the time-dependent energy scaling function $h(t)$ as well as scaling parameter $\lambda$ influence the performance of energy-guided flow matching, we analyze their effects on the D4RL benchmark. Appendix \ref{appendix:lambda} presents results for various values of $\lambda$, specifically $\lambda \in [0.01, 0.05, 0.1, 0.5, 1.0]$ in locomotion and manipulation environments, and $\lambda \in [1.0, 1.5, 2.0, 4.0, 6.0, 8.0]$ in navigation environments, using $h(t)=t^2 / (1-t)$. Additionally, we visualize the impact of different $h(t)$ choices in Figure \ref{fig:h_ablation} for the halfcheetah environments. While performance on the medium-expert dataset remains comparable across variants, the energy-maximizing $h(t)=t^2 /(1-t)$ achieves the best results on the medium and medium-replay datasets. In contrast, the learnable $h_{\theta}(t)$ reaches the lowest performance. Additionally, the result of choosing  $t$ or $t^2$ is almost identical. Overall, this ablation study confirms using $h(t)=t^2 /(1-t)$ as the preferred variant for FlowQ.

\subsection{Limitations}
\label{sec:limitations}
The limitations of the method we propose are the following: (1) From a methodological point of view, one limitation is the assumption $\mathcal{E}_t(\cdot) \approx \lambda(t) \mathcal{E}(\cdot)$, see Section~\ref{sec:flowq}. The approximation of the posterior $\hat p_1(x_1) \propto p_1(x_1) \exp (- \lambda \,\mathcal{E}(x_1))$ is dependent on the choice of $\lambda(t)$, which we highlight with an example in Figure \ref{fig:h_study}. Moreover, the approximation is affecting the performance of FlowQ, hence we include an ablation on the choice of $\lambda(t)$ when using FlowQ in Figure \ref{fig:h_ablation}.
(2) Training time can be reduced for many flow sampling steps as we do not need to backpropagate gradients through actions sampled from the flow model. Still, this does not affect the inference time during evaluations as a full forward pass through the flow path is necessary to sample actions.

\section{Conclusion}
\label{sec:conclusion}
In this paper, we introduce FlowQ, a method that uses energy-guided flow matching to approximate the optimal policy in an offline reinforcement learning setting.
We propose an energy guided probability path and a corresponding Gaussian approximation to define a conditional velocity field that allows us to approximate the optimal policy via flow matching.    
Notably, our method eliminates the need for additional action sampling during policy training, significantly reducing computational overhead compared to conventional flow-based policies.

Our empirical evaluations demonstrate that energy-guided flow matching provides reliable approximations of multimodal target densities in benchmark toy examples. Furthermore, results on the D4RL benchmark show that FlowQ performs on par with other offline reinforcement learning methods, including IQL, CQL, and TD3+BC, as well as diffusion-based approaches such as DiffusionQL, EDP and QIPO. Additionally, we confirm that FlowQ maintains a constant computational cost for policy updates, regardless of the number of time steps, proving its scalability. 
These findings highlight the potential of FlowQ as a practical and efficient alternative for offline reinforcement learning. Future research could explore extending this approach to broader applications and investigating adaptive energy scaling mechanisms to further enhance performance.

\bibliographystyle{rlj}
\bibliography{references}

\newpage
\appendix


\section{Derivations}

\subsection{Conditional Probabilty Path $\hat p_t(x_t \vert x_1)$}
\label{eq:appendix_cond_prob}
We derive the conditional probability path $\hat p_t(x_t \vert x_1)$ from the marginal $\hat p_t(x_t)$ as follows
\begin{align}
    \hat p_t(x_t) &= \int \hat p_t(x_t \vert x_1) \hat p_1(x_1) dx_1\\
    \frac{1}{Z_t}p_t(x_t)\exp(-\lambda(t) \mathcal{E}(x_t)) &= \int \hat p_t(x_t \vert x_1) \frac{1}{Z_1} p_1(x_1) \exp(-\lambda(1) \mathcal{E}(x_1)) dx_1  \\
    p_t(x_t) &=  \int \hat p_t(x_t \vert x_1) \frac{\exp (- \lambda(1) \mathcal{E}(x_1))Z_t}{\exp (- \lambda(t)\mathcal{E}(x_t)) Z_1} dx_1 \text{.}
\end{align}
To ensure that the equation holds, $\hat p_t(x_t \vert x_1)$ needs to satisfy
\begin{equation}
    \hat p_t(x_t \vert x_1) =  \int p_t(x_t \vert x_1) \frac{\exp (- \lambda(t) \mathcal{E}(x_t))Z_1}{\exp (- \lambda(1)\mathcal{E}(x_1)) Z_t} dx_1 \text{.}
\end{equation}

\subsection{First-Order Taylor Expansion}
\label{eq:appendix_taylor}
We want to approximate
\begin{equation}
    \hat p_t(x_t \vert x_1) \propto  p_t(x_t \vert x_1) \exp (- \lambda \mathcal{E}(x_t))
\end{equation}
using the first-order Taylor expansion
\begin{equation}
    \mathcal{E}(x_t) \approx \mathcal{E}(y) + \nabla\mathcal{E}(y)^T(x_t-y)
\end{equation}
around $y$. Therefore, we end up with
\begin{align}
\log \hat p_t(x_t \vert x_1) &\propto  \log p_t(x_t \vert x_1) - \lambda \mathcal{E}(x_t) \\
&\approx -\frac{1}{2\sigma(t)^2} (x_t - \mu)^2 -\lambda \nabla\mathcal{E}(y)^T(x_t-y) + C \\
&= -\frac{x_t^2}{2\sigma(t)^2} + \frac{2x_t\mu}{2\sigma(t)^2} -\frac{\mu^2}{2\sigma(t)^2} - x_t\lambda\nabla\mathcal{E}(y)^T + y\lambda \nabla\mathcal{E}(y)^T \text{.}
\end{align}
Rearranging the terms dependent on $x_t$ we get
\begin{align}
-\frac{x_t^2}{2\sigma(t)^2} +\frac{2x_t\mu}{2\sigma(t)^2} - x_t\lambda\nabla\mathcal{E}(y)^T 
&= -\frac{1}{2\sigma(t)^2} [x_t^2 - 2x_t\mu - 2\sigma(t)^2x_t\lambda\nabla\mathcal{E}(y)^T] \\
&= -\frac{1}{2\sigma(t)^2} [x_t - (\mu - \sigma(t)^2\lambda\nabla\mathcal{E}(y)^T)]^2 - C \\
&= -\frac{1}{2\sigma(t)^2} [x_t - \mu_c]^2
\end{align}
and derive mean and variance as
\begin{equation}
\alpha_c(t, x1) = \alpha(t)x_1 - \lambda \sigma(t)^2\nabla\mathcal{E}(y)
\end{equation}
and
\begin{equation}
    \sigma_c(t)^2 = \sigma(t)^2 \text{.}
\end{equation}

\subsection{Target velocity field $\hat u_t(x_t \vert x_1)$}
\label{eq:target_ut}
We derive the target velocity field $\hat u_t(x_t \vert x_1)$ under the assumption of a linear probability path $\alpha(t)=t$ and $\sigma(t)=1-t$ as follows:
\begin{align}
\hat u_t(x_t \vert x_1) =& \dot \alpha_c(t) + \frac{\dot\sigma_c(t)}{{\sigma_c(t)}}[x_t - \alpha_c(t)]\\
=& \dot\alpha(t)x_1 + \frac{d\lambda \sigma(t)^2\nabla\mathcal{E}(t,y)}{dt} + \frac{\dot\sigma(t)}{{\sigma(t)}}[x_t - \alpha(t)x_1 - \lambda \sigma(t)^2\nabla\mathcal{E}(t,y)] \\
=& x_1  + \frac{tx_1}{{1-t}} - \frac{x_t}{{1-t}} \nonumber \\
 &- \dot\lambda (1-t)^2\nabla\mathcal{E}(y) +2\lambda (1-t) \nabla\mathcal{E}(y) \nonumber \\
 &- \lambda (1-t)^2 \nabla^2\mathcal{E}(y)\dot y  - \lambda (1-t)\nabla\mathcal{E}(y) \\
=& \frac{x_1-x_t}{1-t} + [\lambda - \dot\lambda (1-t)](1-t) \nabla\mathcal{E}(y) - \lambda (1-t)^2 \nabla^2\mathcal{E}(y)\dot y  \text{.}
\end{align}

\section{Implementation Details}
\label{appendix:implementation}
We use the hyper-parameters reported in Table \ref{tab:hp} in our DiffusionQL, (FM)DiffusionQL, EDP and FlowQ experiments. We base our implementation and the choice of hyper-parameters on \cite{wang2023diffusionpoliciesexpressivepolicy}. We implement everything in JAX using Equinox \cite{kidger2021equinox}.

\textbf{FlowQ.}
We run a hyper-parameter search for $\lambda$ over $\{0.01, 0.05, 0.1, 0.5, 1.0\}$ for the locomotion and manipulation environments and over $\{1.0, 1.5, 2.0, 4.0, 6.0, 8.0\}$ for the antmaze environments. We report the results in Table \ref{tab:lambda-locom} and \ref{tab:lambda-navi} and select the highlighted values for the final results reported in Table \ref{tab:d4rl_results} and Table \ref{tab:best_result}. Based on the ablation study in Figure \ref{fig:h_ablation}, we choose $h(t)$ = $\frac{t^2}{1-t}$ for our experiments. As we find, that larger antmaze environments strongly benefit from a larger critic network, we change the critic to $[512, 512, 512, 512]$  with GELU activation functions \cite{gelu} for the antmaze-medium and antmaze-large environments. Moreover, we train the antmaze-large environments for $8\cdot10^5$ gradient steps.

\textbf{(DiffusionQL and EDP.}
We follow the original implementation details in \cite{kang2023efficientdiffusionpoliciesoffline, wang2023diffusionpoliciesexpressivepolicy}.

\textbf{(FM)DiffusionQL.}
We use the same implementation as for DiffusionQL, but replace diffusion with flow matching. The DiffusionQL hyper-parameters from \cite{wang2023diffusionpoliciesexpressivepolicy} work well for the locomotion and manipulation environments as well as for antmaze-umaze and antmaze-umaze-diverse. For the antmaze-medium and antmaze-large environments we run a hyper-parameter search on $\eta$ over $\{0.1, 0.5, 1.0, 1.5, 2.0, 4.0, 6.0, 8.0\}$. We further evaluate different critic architectures, comparing $[256, 256, 256]$ and $[512, 512, 512, 512]$ layers with Tanh activation or GELU activation functions. Still, we could not find a working hyper-parameter setup for these environments.

\begin{table}[h]
\centering
\begin{tabular}{ll}
\toprule
\textbf{Hyper-parameter} & \textbf{Value} \\
\midrule
Gradient steps & $2\cdot10^6$ (locomotion), $1\cdot10^6$ (antmaze, adroit)\\
Batch size & $256$ \\
Optimizer & Adam \citep{kingma2017adammethodstochasticoptimization} \\
Learning rate  & $0.0003$ (default), $0.00003$ (adroit) \\
Discount factor & $0.99$\\
Policy MLP  & $[256, 256, 256]$ with Mish activation \citep{mish} \\ 
Critic MLP  & $[256, 256, 256]$ with Tanh activation (default) \\
Target policy learning rate & 0.005 \\
Discount factor & 0.99 \\
Action candidates & 50 \\
Max Q backup & False (default), True (antmaze) \citep{cql} \\
Reward modification & None (default), CQL-style (antmaze), standardize (adroit)\\
Flow sampling steps & $20$ (FlowQ), $5$ (DiffusionQL, (FM)DiffusionQL), $20$ EDP\\
Flow solver & Euler method \\
\bottomrule
\\
\end{tabular}
\caption{\small Hyper-parameters for DiffusionQL, (FM)DiffusionQL, EDP and FlowQ}
\label{tab:hp}
\end{table}

\section{Computational Resources}
\label{sec:comp_resources}
Each FlowQ experiment requires about 70 minutes on an A100 GPU. To maximize efficiency, we run most computations on GPUs and parallelize wherever possible. However, environment evaluations must run on the CPU, creating a performance bottleneck. Reproducing the full FlowQ benchmark takes roughly 100 GPU-hours, not counting the additional time spent on preliminary implementations, running baselines, ablation studies, or hyper-parameter tuning.

\newpage
\section{Energy scaling hyper-parameter $\lambda$}
\label{appendix:lambda}

In Table \ref{tab:lambda-locom} and \ref{tab:lambda-navi} we present an ablation study investigating how the energy scaling hyper-parameter, $\lambda$, affects FlowQ’s performance. The results indicate that locomotion and manipulation environments require relatively small values of $\lambda$ for optimal performance, while the navigation tasks benefit from larger $\lambda$ values.

\begin{table}[h]
\centering
\begin{tabular}{llllll}
\toprule
$\lambda$                       & 0.01           & 0.05           & 0.10          & 0.50  & 1.00          \\
\midrule
halfcheetah-medium-v2        & 48.9 $\pm$ 0.3            & 50.2 $\pm$ 0.3           & 51.1 $\pm$ 0.4          & 54.3 $\pm$ 0.6  & \textbf{56.8 $\pm$ 0.8} \\
halfcheetah-medium-replay-v2 & 45.5 $\pm$ 0.3           & 46.3 $\pm$ 0.3           & 47.1 $\pm$ 0.3          & 49.7 $\pm$ 0.5  & \textbf{50.3 $\pm$ 0.9} \\
halfcheetah-medium-expert-v2 & \textbf{89.6 $\pm$ 2.3}  & 88.3 $\pm$ 5.0           & 84.6 $\pm$ 6.0          & 47.4 $\pm$ 11.9  & 49.3 $\pm$ 7.9          \\
hopper-medium-v2             & 67.8 $\pm$ 7.1           & 76.4 $\pm$ 8.4           & \textbf{88.9 $\pm$ 6.4} & 82.8 $\pm$ 12.1  & 68.0 $\pm$ 14.9          \\
hopper-medium-replay-v2      & 93.3 $\pm$ 9.5           & \textbf{99.2 $\pm$ 1.4}  & 99.1 $\pm$ 1.9          & 45.4 $\pm$ 20.4  & 29.2 $\pm$ 6.9          \\
hopper-medium-expert-v2      & \textbf{103.5 $\pm$ 7.0} & 99.9 $\pm$ 9.1           & 97.1 $\pm$ 9.1         & 69.3 $\pm$ 18.8  & 0.0 $\pm$ 0           \\
walker2d-medium-v2           & 84.1 $\pm$ 0.3           & \textbf{84.9 $\pm$ 0.4}  & 84.4 $\pm$ 2.5          & 39.3 $\pm$ 6.4  & 23.5 $\pm$ 10.6          \\
walker2d-medium-replay-v2    & 83.8 $\pm$ 5.4           & \textbf{88.5 $\pm$ 2.9}  & 84.6 $\pm$ 4.9          & 49.1 $\pm$ 17.9  & 56.0 $\pm$ 11.6          \\
walker2d-medium-expert-v2    & 109.0 $\pm$ 0.2          & \textbf{109.1 $\pm$ 0.2} & 109.1 $\pm$ 0.2         & 108.1 $\pm$ 2.9 & 49.3 $\pm$ 28.5          \\
\midrule
pen-human-v1  & \textbf{69.5 $\pm$ 19.0} & 62.8 $\pm$ 18.8  & 48.4 $\pm$ 17.7  & 26.4 $\pm$ 17.9  & 33.6 $\pm$ 20.1          \\
pen-cloned-v1 & 65.0 $\pm$ 20.5          & 69.1 $\pm$ 17.5  & 71.9 $\pm$ 22.0  & 86.1 $\pm$ 20.3  & \textbf{91.6 $\pm$ 23.5}
\\
\bottomrule
\\
\end{tabular}
\caption{\small $\lambda$ hyper-parameter for the D4RL locomotion and manipulation environments.}
\label{tab:lambda-locom}
\end{table}

\begin{table}[h]
\centering
\begin{tabular}{lllllll}
\toprule
$\lambda$                    & 1.00 & 1.50         & 2.00          & 4.00  & 6.00          & 8.00          \\
\midrule
antmaze-umaze-v2          & 60.8 $\pm$ 15.5 & 83.0 $\pm$ 13.0        & \textbf{92.8 $\pm$ 7.9} & 70.2 $\pm$ 22.1  & - & -            \\
antmaze-umaze-diverse-v2  &  58.5 $\pm$ 13.7 & 64.0 $\pm$ 13.1        & 57.8 $\pm$ 13.5         & \textbf{65.2 $\pm$ 13.3} & - & -   \\
antmaze-medium-play-v2     & 33.0 $\pm$ 18.5 & 83.4 $\pm$ 17.4 & \textbf{84.8 $\pm$ 7.9} & 81.9 $\pm$ 8.7        & 81.8 $\pm$ 7.2         & 73.0 $\pm$ 10.6   \\
antmaze-medium-diverse-v2 & 2.3 $\pm$ 4.0 &  \textbf{44.7 $\pm$ 22.8}  & 41.4 $\pm$ 26.6  & 19.5 $\pm$ 18.4          & 11.5 $\pm$ 12.0 & 14.3 $\pm$ 14.1          \\
antmaze-large-play-v2     & - & -  & 25.6 $\pm$ 18.9    & \textbf{72.4 $\pm$ 11.5} & 62.6 $\pm$ 18.6 & 61.9 $\pm$ 20.9  \\
antmaze-large-diverse-v2  & - & -  & 13.7 $\pm$ 9.0  & 40.4 $\pm$ 17.4 & 49.1 $\pm$ 18.6           & \textbf{55.8 $\pm$ 16.8}       \\
\bottomrule
\\
\end{tabular}
\caption{\small $\lambda$ hyper-parameter for the D4RL navigation environments.}
\label{tab:lambda-navi}
\end{table}

\newpage
\section{Individual training runs of FlowQ}
In this section we provide the individual training runs of FlowQ which are used for generating the results in Table \ref{tab:d4rl_results}.
\begin{figure}[ht]
    \begin{center}
        \includegraphics[width=0.7\columnwidth]{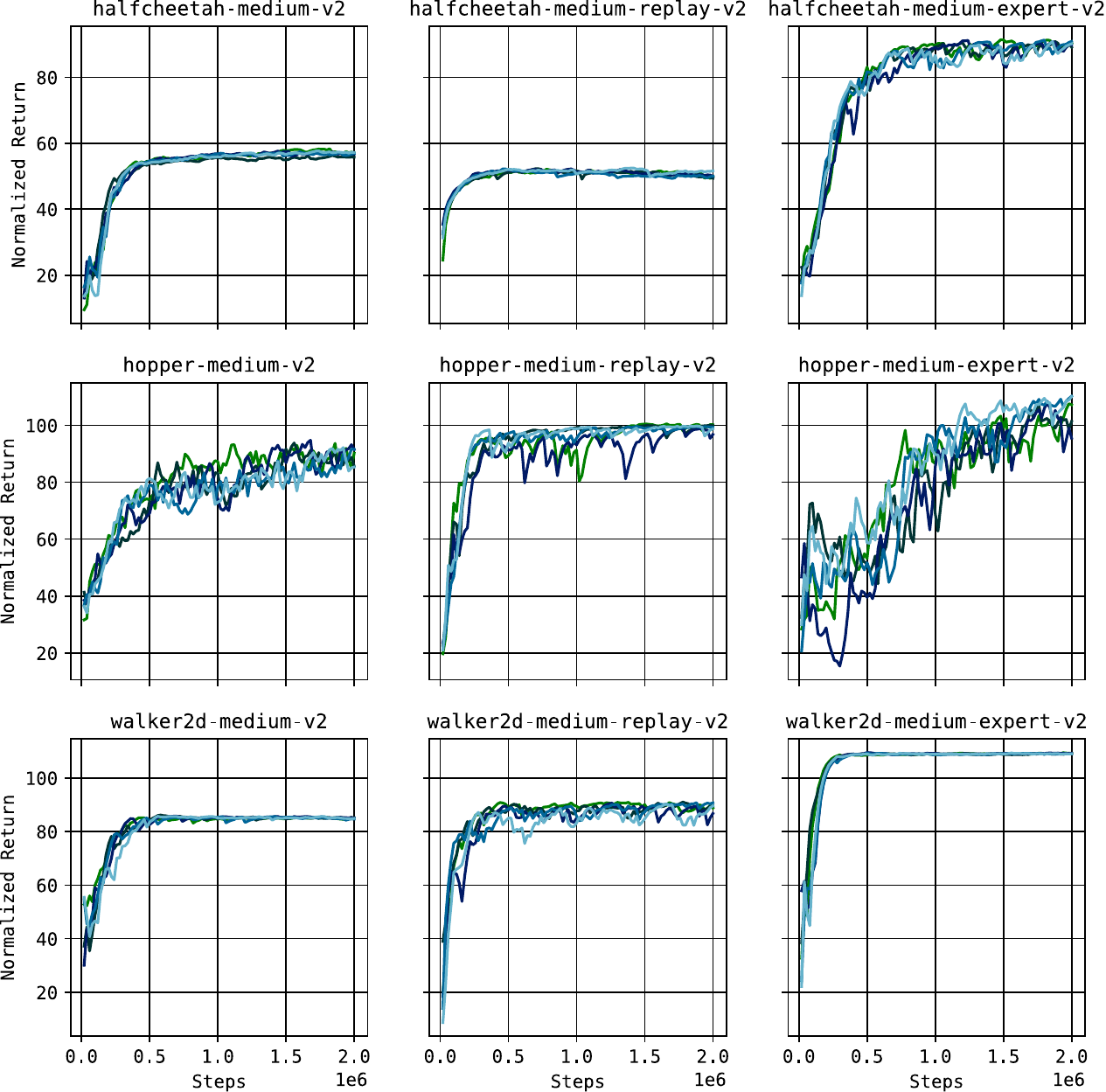}
    \end{center}
    \caption{\small Normalized return of FlowQ for the D4RL locomotion environments using 5 seeds.}
    \label{fig:flowq_locomotion_results}
\end{figure}
\begin{figure}[ht]
    \begin{center}
        \includegraphics[width=0.7\columnwidth]{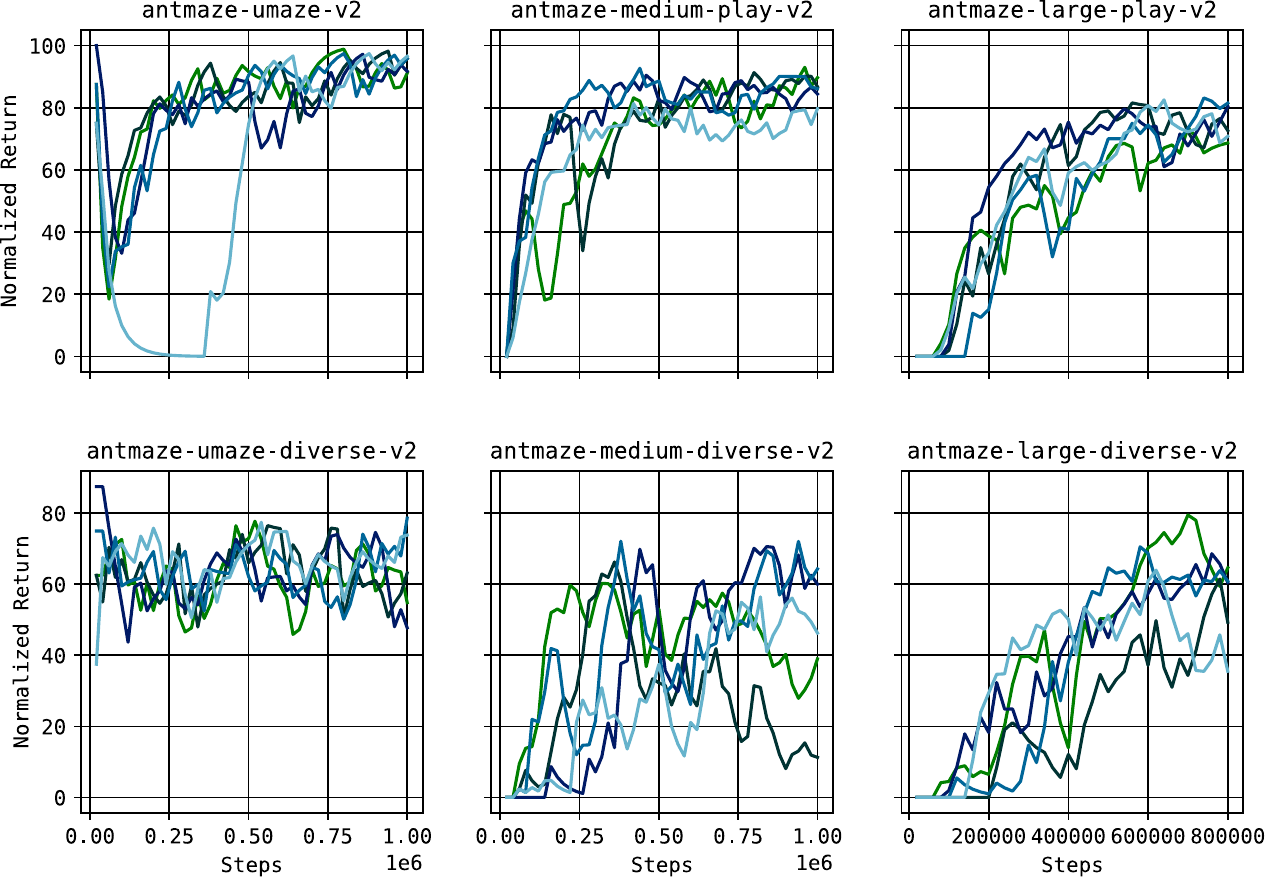}
    \end{center}
    \caption{\small Normalized return of FlowQ for the D4RL adroit environments using 5 seeds.}
    \label{fig:flowq_antmaze_results}
\end{figure}
\begin{figure}[ht]
    \begin{center}
        \includegraphics[width=0.5\columnwidth]{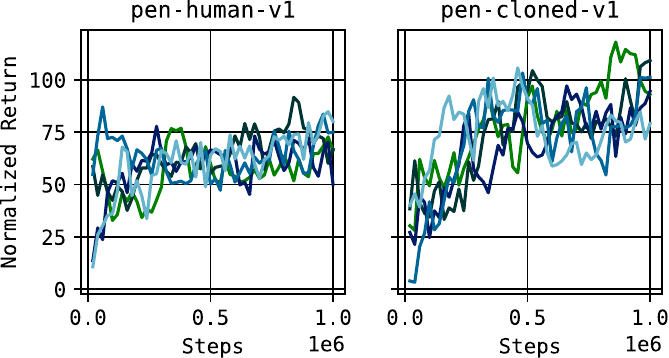}
    \end{center}
    \caption{\small Normalized return of FlowQ for the D4RL adroit environments using 5 seeds.}
    \label{fig:flowq_adroit_results}
\end{figure}

\section{Results when using the \textit{best result during training} evaluation method}
In addition to the \textit{running average at training} results reported in Table \ref{tab:d4rl_results}, we also include, in Table \ref{tab:best_result}, the \textit{best result during training} metrics following the protocol of \cite{zhang2025energyweighted}. We want to highlight, however, that selecting the single best training outcome is not a robust performance indicator. Therefore, the running‐average at the end of training approach is the preferred evaluation method.

\begin{table}[ht]
\centering
\resizebox{0.5\textwidth}{!}{%
\begin{tabular}{lccc}
\toprule
Dataset                   & DiffusionQL & QIPO    & FlowQ(ours)\textsuperscript{*} \\
halfcheetah-medium        & 51.5 & 54.2   & \textbf{58.3 $\pm$ 0.6}  \\
halfcheetah-medium-replay & 48.3 & 48.0   & \textbf{53.1 $\pm$ 0.2}  \\
halfcheetah-medium-expert & 97.3 & \textbf{94.5}   & 92.7 $\pm$ 0.6  \\
hopper-medium             & 96.6 & 94.1   & \textbf{100.4 $\pm$ 1.7}   \\
hopper-medium-replay      & 102.0 & 101.3  & \textbf{101.0 $\pm$ 0.4}  \\
hopper-medium-expert      & 112.3 & 108.0  & \textbf{113.0 $\pm$ 0.5} \\
walker2d-medium           & 87.3 & 87.6   & \textbf{88.5 $\pm$ 1.0}  \\
walker2d-medium-replay    & 98.0 & 78.6   & \textbf{93.9 $\pm$ 0.7}  \\
walker2d-medium-expert    & 111.2 & \textbf{110.9}  & 110.8 $\pm$ 0.3 \\
\rowcolor[HTML]{EFEFEF} 
average                   & 89.3 & 86.3   & \textbf{90.2}   \\
antmaze-umaze             & 96.0 & 93.6   & \textbf{100.0 $\pm$ 0.0}  \\
antmaze-umaze-diverse     & 84.0 & 76.1   & \textbf{100.0 $\pm$ 0.0}  \\
antmaze-medium-play       & 79.8 & 80.0   & \textbf{100.0 $\pm$ 0.0}  \\
antmaze-medium-diverse    & 82.0 & \textbf{86.4}   & 85.0 $\pm$ 4.5  \\
antmaze-large-play        & 49.0 & 55.5   & \textbf{91.0 $\pm$ 2.0}   \\
antmaze-large-diverse     & 61.7 & 32.1   & \textbf{84.0 $\pm$ 5.8}   \\
\rowcolor[HTML]{EFEFEF} 
average                   & 75.4 & 72.0   & \textbf{93.3}      \\
pen-human                 & 75.7 & –       & \textbf{121.3 $\pm$ 12.4}  \\
pen-cloned                & 60.8 & –       & \textbf{144.5 $\pm$ 12.0}  \\
\rowcolor[HTML]{EFEFEF} 
average                   & 68.3 & –       & \textbf{132.9} \\
\bottomrule
\\
\end{tabular}%
}
\caption{\small Averaged normalized score on the D4RL benchmark. We report results (mean and standard deviation) using the \textit{best result during training} method as used in \cite{zhang2025energyweighted} and use five seeds for each experiment. We report the results for DiffusionQL from \cite{wang2023diffusionpoliciesexpressivepolicy} for online model selection, which is a different evaluation protocol. The results for methods annotated by * come from our own implementation.}
\label{tab:best_result}
\end{table}

\end{document}